\documentclass[letterpaper, 10 pt, journal, twoside]{IEEEtran}

\IEEEoverridecommandlockouts                              

\usepackage{graphics} 
\usepackage{epsfig} 
\usepackage{empheq}
\usepackage{times} 
\usepackage{amsmath} 
\usepackage{amssymb}  
\usepackage{bm}
\usepackage{color}
\usepackage{bbding} 
\usepackage{cite}
\usepackage{diagbox}
\usepackage[linesnumbered,ruled]{algorithm2e}
\usepackage{ulem} 
\usepackage{hyperref}
\hypersetup{
colorlinks=true,
linkcolor=blue,
filecolor=blue,      
urlcolor=blue,
citecolor=cyan,
}
\usepackage{float}
\usepackage{booktabs}
\usepackage{multirow}
\usepackage{mathrsfs}
\usepackage[utf8]{inputenc}
\usepackage{subfigure}
\usepackage{graphicx}
\usepackage{pifont}
\usepackage{threeparttable}
\usepackage{caption}
\usepackage{setspace}
\newcounter{RNum}

\newcommand{\cmark}{\ding{51}}%
\newcommand{\xmark}{\ding{55}}%

\newcommand{\hl}[1]{\textcolor{black}{#1}}

\begin{document}

\author{Yiming Li$^{1}$, Dekun Ma$^{1}$, Ziyan An$^{1}$, Zixun Wang$^{1}$, Yiqi Zhong$^{2}$,  Siheng Chen\textsuperscript{3, \ding{41}}, and Chen Feng\textsuperscript{1, \ding{41}}
\\
\thanks{Manuscript received February 24, 2022; Revised June 7, 2022; Accepted June 30, 2022.}
\thanks{This paper was recommended for publication by Editor Cesar Cadena upon evaluation of the Associate Editor and Reviewers' comments. This work was supported by the NSF CPS program under CMMI-1932187, the National Natural Science Foundation of China under Grant 62171276, the Science and Technology Commission of Shanghai Municipal under Grant 21511100900 and CALT Grant 2021-01.}
\thanks{\ding{41} Corresponding author.}
\thanks{$^{1}$Yiming Li, Dekun Ma, Ziyan An, Zixun Wang, and Chen Feng are with are with New York University,
Brooklyn, NY 11201, USA  {\tt\small cfeng@nyu.edu}}%
\thanks{$^{2}$Yiqi Zhong is with University of Southern California, Los Angeles, USA {\tt\small yiqizhon@usc.edu}}%
\thanks{$^{3}$Siheng Chen is with Cooperative Medianet Innovation Center, Shanghai Jiao Tong University and Shanghai AI Laboratory, Shanghai, China {\tt\small sihengc@sjtu.edu.cn}}%
\thanks{Digital Object Identifier (DOI): see top of this page.}
}

\title{V2X-Sim: Multi-Agent Collaborative Perception \\Dataset and Benchmark for Autonomous Driving
}

\markboth{IEEE Robotics and Automation Letters. Preprint Version. JULY, 2022}
{LI \MakeLowercase{\textit{et al.}}: V2X-Sim: Multi-Agent Collaborative Perception Dataset and Benchmark for Autonomous Driving}

\maketitle

\begin{abstract}
	Vehicle-to-everything (V2X) communication techniques enable the collaboration between vehicles and many other entities in the neighboring environment, which could fundamentally improve the perception system for autonomous driving. However, the lack of a public dataset significantly restricts the research progress of collaborative perception. To fill this gap, we present V2X-Sim, a comprehensive simulated multi-agent perception dataset for V2X-aided autonomous driving. V2X-Sim provides: (1) \hl{multi-agent} sensor recordings from the road-side unit (RSU) and multiple vehicles that enable collaborative perception, (2) multi-modality sensor streams that facilitate multi-modality perception, and (3) diverse ground truths that support various perception tasks. Meanwhile, we build an open-source testbed and provide a benchmark for the state-of-the-art collaborative perception algorithms on three tasks, including detection, tracking and segmentation. V2X-Sim seeks to stimulate collaborative perception research for autonomous driving before realistic datasets become widely available.  Our dataset and code are available at \url{https://ai4ce.github.io/V2X-Sim/}.
\end{abstract}


\section{Introduction}\label{sec:intro}

\IEEEPARstart{P}{erception} is a fundamental \hl{capability} for autonomous vehicles, which allows them to represent, identify, and interpret sensory input for understanding the complex surroundings. In literature, single-vehicle perception has been intensively studied thanks to the well-established driving datasets~\cite{Caesar2020nuScenesAM,Geiger2012AreWR,Sun2020ScalabilityIP}, and researchers have proposed various algorithms to deal with different downstream tasks~\cite{arnold2019survey,marvasti2021deep,Minaee2021ImageSU}. 

Despite recent advances in single-vehicle perception, the individual viewpoint often results in degraded perception in long-range or occluded areas. A promising solution to this problem is through vehicle-to-everything (V2X)~\cite{machardy2018v2x}, a cutting-edge communication technology that enables dialogue between a vehicle and other entities, including vehicle-to-vehicle (V2V) and vehicle-to-infrastructure (V2I). With the aid of V2X communication, we are able to upgrade single-vehicle perception to~{collaborative perception}, which introduces more viewpoints to help autonomous vehicles see further, better and even see through occlusion, thereby fundamentally enhancing the capability of perception.

Collaborative perception naturally draws on communication and perception. Its development requires expertise from both communities. Recently, the communication community has made enormous efforts to promote such a study~\cite{muhammad2018survey,hasan2020securing,mannoni2019comparison}; however, only a few works have been proposed from the perspective of perception~\cite{Chen2019CooperCP,Wang2020V2VNetVC, Li_2021_NeurIPS,yuan2021comap,yuan2022keypoints}.
One major reason for this is the lack of well-designed and organized collaborative perception datasets. Due to the immaturity of V2X and the cost of simultaneously operating multiple autonomous vehicles, it is very expensive and laborious to build such a real dataset for research communities. Therefore, \textit{we synthesize a comprehensive and publicly available dataset, named as V2X-Sim, to advance the study of collaborative perception for V2X-communication-aided autonomous driving.}

 \begin{figure}[t]
	\centering	\includegraphics[width=0.95 \columnwidth]{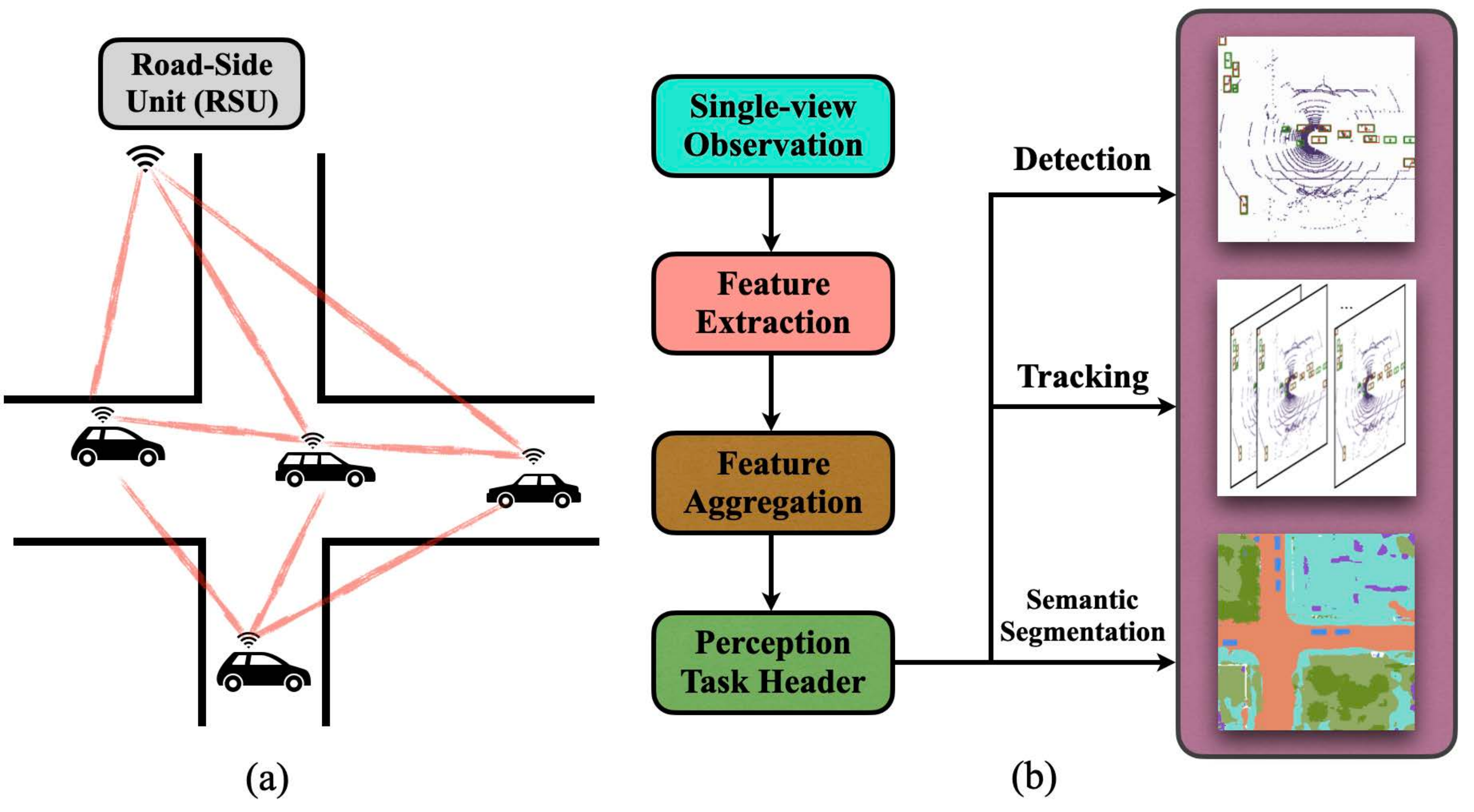}
	\captionsetup{font={scriptsize}}
	\caption{ (a) \textbf{Intersection} for vehicle-to-everything (V2X) communication. (b) \textbf{Workflow} of multi-agent collaborative perception with intermediate-/feature-based strategy. We benchmark collaborative object detection, multi-object tracking, and semantic segmentation in the bird's eye view (BEV).}
	\label{fig:1}
	\vspace{-0.5cm}
\end{figure}

To generate V2X-Sim, we employ SUMO~\cite{krajzewicz2012recent}, a micro-traffic simulation, to produce numerically-realistic traffic flow, and CARLA~\cite{Dosovitskiy17}, a widely-used open-source simulator for autonomous driving research, to retrieve well-synchronized sensor streams from multiple vehicles as well as the road-side unit (RSU). Meanwhile, multi-modality sensor streams of different entities are recorded to enable cross-modality perception. In addition, diverse annotations including bounding boxes, vehicle trajectories, and semantic labels are provided to facilitate various downstream tasks. To better serve multi-agent, multi-modality, and multi-task perception research for autonomous driving, we further provide a benchmark for three crucial perception tasks (collaborative detection, tracking, and segmentation) on the proposed dataset using the state-of-the-art collaboration strategies~\cite{Li_2021_NeurIPS,Wang2020V2VNetVC,Liu2020When2comMP,Liu2020Who2comCP}. In summary, our contributions are two-fold: 
\begin{itemize}
     \item We propose V2X-Sim, a comprehensive V2X perception dataset for autonomous driving, to support multi-agent multi-modality multi-task perception research.
    \item We create an open-source testbed for collaborative perception methods, and provide a benchmark on three tasks to encourage more research in this field. 
\end{itemize}

\begin{table}[t]
\captionsetup{font={scriptsize, sc, stretch=1.3}, justification=centering, labelsep=newline}
\caption{
\label{tab:datasetcomp} Comparison of collaborative perception datasets for autonomous driving. There are no public datasets which support both V2V and V2I research: multi-agent data are either generated by simulators~\cite{xu2021opv2v,Wang2020V2VNetVC} or created by selecting consecutive frames from single-agent real datasets~\cite{Chen2019CooperCP,Xiao2018MultimediaFA,Maalej2017VANETsMA}. \hl{Several works collect data from multiple infrastructure sensors:} \cite{arnold2020cooperative} in simulation,\cite{howe2021weakly,yu2022dairv2x} in real world.  Our dataset is the first public multi-agent multi-modality dataset which supports different collaborative perception tasks.}
\tiny
  \centering
   \setlength{\tabcolsep}{0.1mm}{
  \begin{tabular}{@{}c|c|c|cccc|ccc|c@{}}
      \toprule
    \multirow{2}{*}{\bf Dataset} & \multirow{2}{*}{\bf Scenario}& \multirow{2}{*}{\bf Source} & \multicolumn{4}{c|}{\bf Sensor} &  \multicolumn{3}{c|}{\bf Tasks}& \multirow{2}{*}{\bf Public} \\
      &&& {\bf RGB} & {\bf Depth} &  {\bf LiDAR} &  {\bf IMU/GPS} & {\bf Det.} & {\bf Track.} & {\bf Seg.}& 
      \\
    \midrule
    V2V-Sim~\cite{Wang2020V2VNetVC} & V2V & LiDARsim~\cite{Manivasagam2020LiDARsimRL} & \xmark & \xmark & \cmark & \xmark & \cmark &
    {\cmark} & \xmark  &\xmark \\
    Cooper~\cite{Chen2019CooperCP} & V2V & KITTI~\cite{Geiger2012AreWR} & \xmark & \xmark & \cmark & \cmark & \cmark &
    {\cmark} & \xmark  &\xmark \\
    MFSL~\cite{Xiao2018MultimediaFA} & V2V & KITTI~\cite{Geiger2012AreWR} & \cmark & \xmark & \xmark & \cmark & \xmark &
    \xmark & \cmark  &\xmark \\
    VANETs~\cite{Maalej2017VANETsMA} & V2V & KITTI~\cite{Geiger2012AreWR} & \cmark & \xmark & \cmark & \cmark & \cmark &
    \xmark & \xmark  &\xmark \\ 
    OPV2V~\cite{xu2021opv2v} & V2V & OpenCDA~\cite{xu2021opencda} \& CARLA~\cite{Dosovitskiy17} & \cmark & \xmark & \cmark & \cmark & \cmark &
    {\cmark} & \xmark  &\cmark \\ 
    CODD~\cite{arnold2021fastreg} & V2V &  CARLA~\cite{Dosovitskiy17} & \xmark & \xmark & \cmark & \xmark & \cmark &
    {\cmark} & \xmark  &\cmark \\ 
    \hl{Cooper (inf)}~\cite{arnold2020cooperative}  & V2I & CARLA~\cite{Dosovitskiy17} & \cmark & \cmark & \xmark & \xmark & \cmark & \cmark & \xmark  &\cmark\\ 
    \hl{WIBAM}~\cite{howe2021weakly} & V2I & Real-world & \cmark & \xmark & \xmark & \xmark & \cmark &
    \cmark & \xmark  &\cmark \\ 
    \hl{DAIR-V2X}~\cite{yu2022dairv2x} & V2I & Real-world & \cmark & \xmark & \cmark & \cmark & \cmark &
    \xmark & \xmark  &\cmark \\ 
    \midrule
    V2X-Sim (\textbf{Ours}) & V2V\&V2I & CARLA~\cite{Dosovitskiy17} \& SUMO~\cite{krajzewicz2012recent} & \cmark & \cmark & \cmark & \cmark & \cmark &
    \cmark & \cmark &\cmark \\
    \bottomrule
    \end{tabular}
    }
   \vspace{-4mm}
\end{table}

\section{Related Work}\label{sec:related}
\textbf{Autonomous driving dataset.} 
Since the pioneering dataset KITTI~\cite{Geiger2012AreWR} was released, the autonomous driving community has been trying to increase the dataset comprehensiveness in terms of driving scenarios, sensor modalities, and data annotations. Regarding driving scenarios, current datasets cover crowded urban scenes~\cite{Patil2019TheHD}, adverse weather conditions~\cite{Pitropov2021CanadianAD}, night scenes~\cite{Pham2020A3DDT}, and multiple cities~\cite{Caesar2020nuScenesAM} to enrich the data distribution. As for sensor modalities, nuScenes~\cite{Caesar2020nuScenesAM} collects data with Radar, RGB cameras, and LiDAR in a 360$^{\circ}$ viewpoint; WoodScape~\cite{Yogamani2019WoodScapeAM} captures data with fisheye cameras. Regarding data annotations, semantic labels in both images~\cite{Huang2018TheAD, Cordts2016TheCD, Ros2016TheSD, Neuhold2017TheMV} and point clouds~\cite{Behley2019SemanticKITTIAD, Hu2020TowardsSS} are provided to enable semantic segmentation; 2D/3D box trajectories are offered~\cite{Chang2019Argoverse3T,Ettinger2021LargeSI} to facilitate tracking and prediction. In summary, most real datasets emphasize the data comprehensiveness in single-vehicle situations, but ignore the multi-vehicle scenarios.

\textbf{V2X system and dataset.} By sharing information with other vehicles or the RSU, V2X mitigates the shortcomings of single-vehicle perception and planning such as the limited sensing range and frequent occlusion~\cite{machardy2018v2x}. Previous research~\cite{Jia2016EnhancedCC} has developed an enhanced cooperative microscopic traffic model in V2X scenarios, and studied the effect of V2X in traffic disturbance scenarios. \cite{Kim2015MultivehicleCD} proposes a multi-modal cooperative perception system that provides see-through, lifted-seat, satellite and all-around views to drivers. \hl{More recently, COOPER}~\cite{Chen2019CooperCP} \hl{investigates raw-data level collaborative perception to improve the detection capability for autonomous driving. V2VNet}~\cite{Wang2020V2VNetVC}
\hl{proposes intermediate-feature level collaboration to promote the vehicle's perception and prediction capability. Several works use multiple infrastructure sensors to jointly perceive the environment and employ output-level collaboration with vehicle-to-infrastructure communication}~\cite{arnold2020cooperative, howe2021weakly}. As for the datasets, ~\cite{Chen2019CooperCP,Xiao2018MultimediaFA,Maalej2017VANETsMA} simulate the V2V scenarios with different frames from KITTI~\cite{Geiger2012AreWR}. Yet, these datasets are unrealistic multi-agent datasets for the measurements are not captured by different viewpoints. Some other works use a platoon strategy for data capture~\cite{Rawashdeh2018CollaborativeAD, Chen2015DSRCAR}, but they are biased because the observations were highly correlated with each other.
The most relevant work is V2V-Sim~\cite{Wang2020V2VNetVC}, which is based on a high-quality LiDAR simulator~\cite{Manivasagam2020LiDARsimRL}. Unfortunately, V2V-Sim does not include the V2I scenario and is not publicly available. Moreover, OPV2V~\cite{xu2021opv2v} and CODD~\cite{arnold2021fastreg} only support the detection task in the V2V scenario. Existing collaborative perception datasets are summarized in Table~\ref{tab:datasetcomp}: \textit{V2X-Sim\footnote{This work extends the LiDAR-based V2V data in our previous work~\cite{Li_2021_NeurIPS} with more modalities, scenarios and downstream tasks.} is currently the most comprehensive one with multi-agent multi-modality sensory streams in both V2V and V2I scenarios, and can support various downstream tasks such as multi-agent collaborative detection, tracking, and semantic segmentation.}

\begin{figure}[t]
\centering
    \begin{minipage}{.5\linewidth}
    \centering
    \captionsetup{font={scriptsize, sc, stretch=1.3}, justification=centering, labelsep=newline}
    \captionof{table}{Sensor suite of vehicle (\textbf{V}) and intersection (\textbf{I}). }
        \scriptsize
        \vspace{-2mm}
        \begin{center}
        \resizebox{1\textwidth}{!}{%
            \setlength{\tabcolsep}{1mm}{
        \begin{tabular}[H]{ p{2cm}|p{5cm}}
         \toprule
         Sensor & Description\\
         \midrule
         \textbf{V:} 6 $\times$ RGB camera \textbf{I:} 4 $\times$ RGB camera  & Each vehicle is equipped with 6 cameras. Each camera has a FoV of 70$^{\circ}$, except for the back camera that has a FoV of 110$^{\circ}$. Each RSU has 4 cameras looking diagonally downward at 35$^{\circ}$ with a 70$^{\circ}$ FoV. 
         The image size is 1600$\times$900.  \\
         \midrule
         \textbf{V:} 6 $\times$ Depth camera \textbf{I:} 4 $\times$ Depth camera & Each vehicle has 6 depth cameras with  the same setting as RGB cameras. Each RSU has 4 depth cameras with  the same setting as its RGB cameras. \\
          \midrule
         \textbf{V:} 6 $\times$ Semantic camera \textbf{I:} 4 $\times$ Semantic camera  & Each vehicle has 6 semantic segmentation cameras with the same setting as RGB cameras. Each RSU has 4 semantic segmentation cameras with the same setting as its RGB cameras. \\
          \midrule
         \textbf{V\&I:} 1 $\times$ BEV semantic camera & Each vehicle and RSU has one BEV semantic camera at the top, looking downward. Both the raw images (semantic tags encoded in the red channel) and the converted colored images are provided. The image size is 900$\times$900. \\
          \midrule
         \textbf{V\&I:} 1 $\times$ LiDAR and Semantic LiDAR &  We attach one LiDAR and one semantic LiDAR on top of the ego vehicle and the intersection center. Specs: 32 channels, 70m max range, 250,000 points per second, 20 Hz rotation frequency.\\
         \bottomrule
        \end{tabular}}}
                
        \label{tab:setup}
        \end{center}
    \vspace{-4mm}
    \end{minipage}~~
    \begin{minipage}{.47\linewidth}
    \centering
  \includegraphics[width=1.0\textwidth]{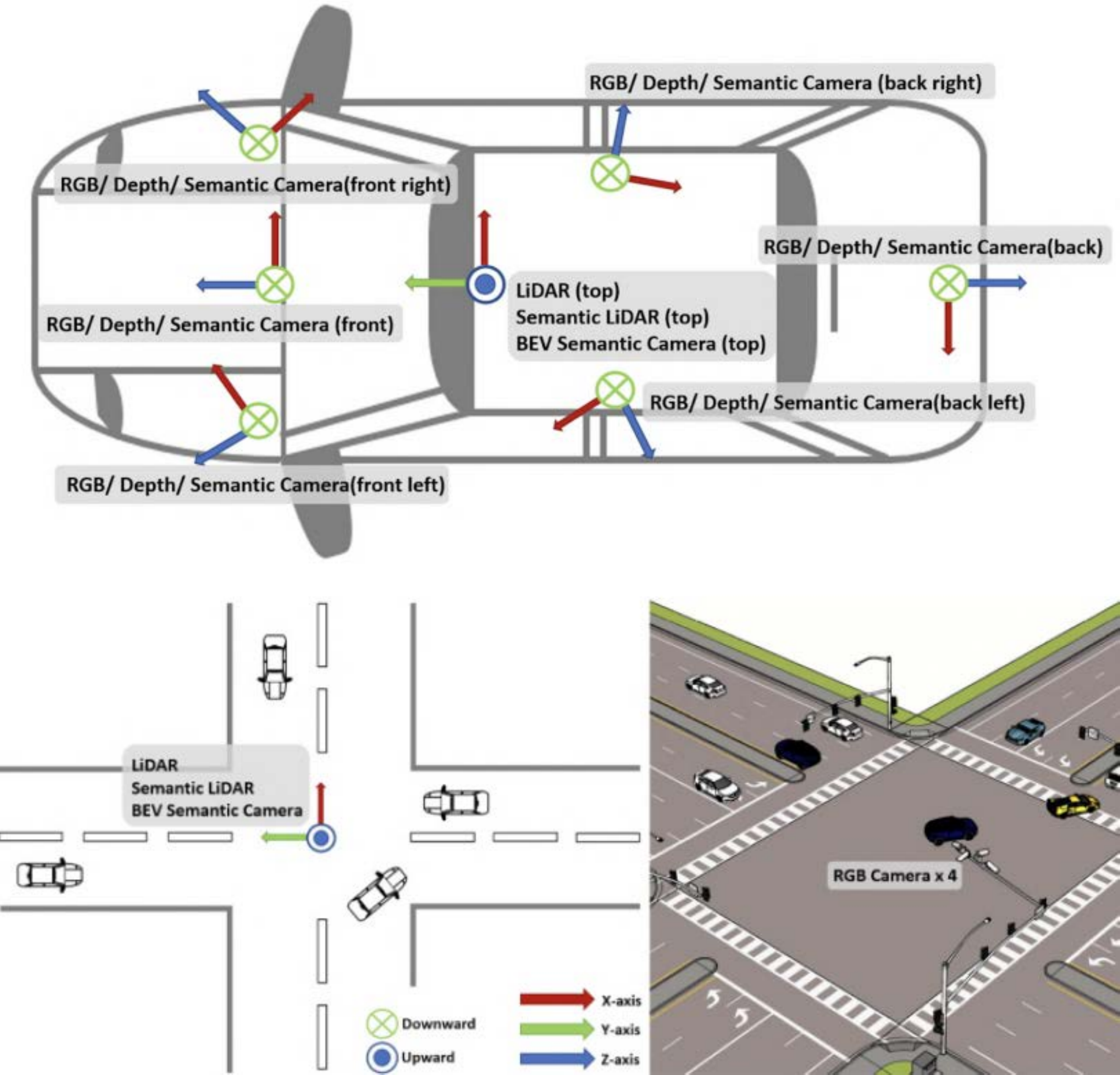}
  \captionsetup{font={scriptsize}}
  \caption{Sensor layout and coordinate systems.}
  \label{fig:setup}
    \end{minipage}~~
\vspace{-2mm}
\end{figure}

\begin{figure*}[t]
\begin{center}
\includegraphics[width=0.95\textwidth]{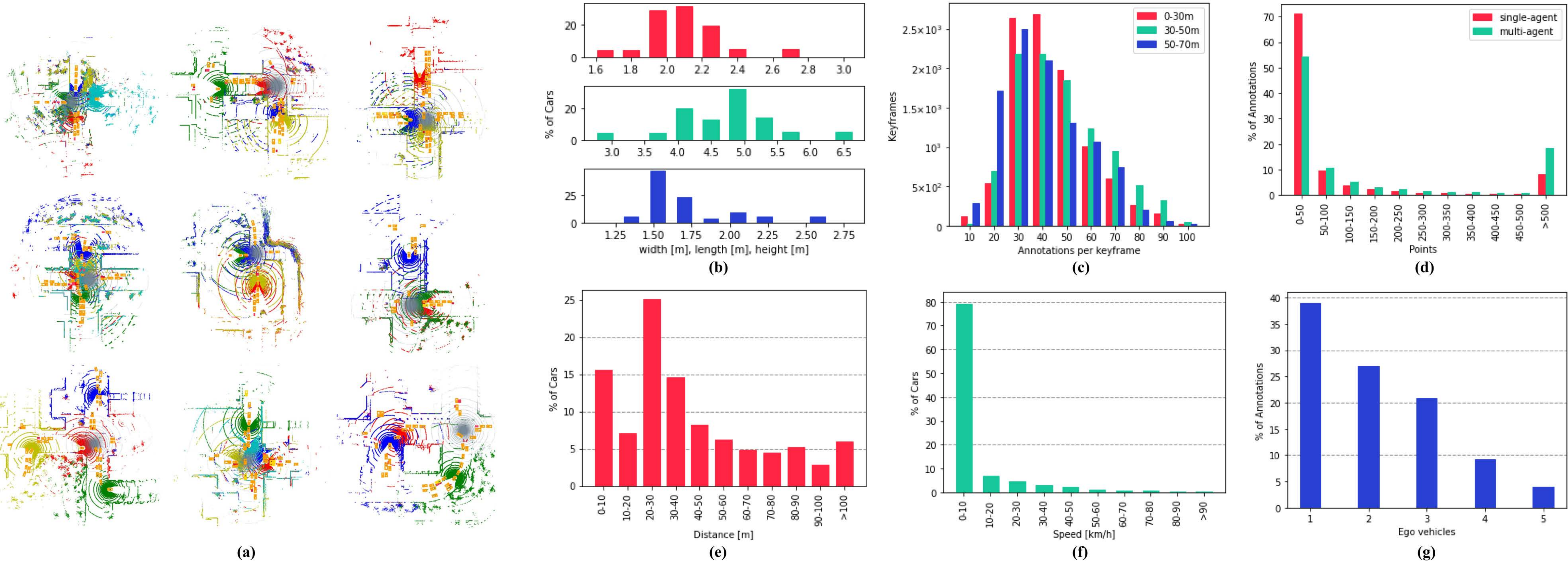}
\vspace{-1mm}
\captionsetup{font={scriptsize}}
\caption{\hl{(a) Visualizations of the bird's eye view point cloud from different scenes. Gray denotes the point cloud captured by the RSU LiDAR. Each color (except for gray) represents a vehicle, and the orange boxes denote the vehicles in the scene. (b) Statistics for car bounding box sizes. (c) Counts for annotations per keyframe where the annotated vehicles are presented within 0-30m, 30-50m, and 50-70m of the ego vehicles. (d) Counts for LiDAR points per annotation. (e) Statistics of the distance between every two ego vehicles for all frames. (f) Speed of cars located within 70m from ego vehicles. (g) Percentage of annotated vehicles observed by 1-5 ego vehicles.}}
\label{fig:scene}
\end{center}
\vspace{-8mm}
\end{figure*}
\section{V2X-Sim Dataset}
V2X-Sim could enable more research on the collaboration strategy among vehicles to achieve a more robust perception. This could fundamentally benefit autonomous driving, intelligent transportation systems, smart cities, \textit{etc.}.
\subsection{Sensor suite on vehicles and RSU}
Multi-modality data is essential for robust perception. To ensure the comprehensiveness of our dataset, we equip each vehicle with a sensor suite composed of RGB cameras, LiDAR, GPS, IMU, and RSU with RGB cameras and LiDAR. 

\textbf{Sensor configuration.} On both the vehicle and RSU, the camera and LiDAR cover 360$^{\circ}$ horizontally to enable full-view perception. Specifically, each vehicle carries six RGB cameras following the nuScenes configuration~\cite{Caesar2020nuScenesAM}; the RSU is equipped with four RGB cameras toward four directions at the crossroad. We employ depth camera, semantic segmentation camera, semantic LiDAR, and BEV semantic segmentation camera in CARLA to obtain the corresponding ground-truth for each RGB camera. Note that the BEV semantic segmentation camera is based on orthogonal projection while the ego-vehicle semantic segmentation camera uses perspective projection. Table~\ref{tab:setup} summarizes the detailed sensor specification. 

\textbf{Sensor layout and coordinate system.}
The overall sensor layout and coordinate system is shown in Fig.~\ref{fig:setup}. The BEV semantic camera shares the same x, y position with LiDAR yet is placed higher to ensure a certain size of field of view. Note that we invert the y-axis in CARLA and use a right-hand coordinate system following nuScenes~\cite{Caesar2020nuScenesAM}. 

\textbf{Diverse annotations.}
To assist downstream tasks including detection, tracking and semantic segmentation, we provide various annotations such as 3D bounding boxes, pixel-wise and point-wise semantic labels. Each box is defined by the location of its center in x, y, z coordinates, and its width, length, and height. In total, there are 23 categories such as the pedestrian, building, ground, etc. In addition, precise depth values are provided for depth estimation.

\subsection{CARLA-SUMO co-simulation}
We consider it a realistic V2X scenario when multiple vehicles with their own routes are simultaneously located in the same intersection. \hl{ Each intersection is also equipped with one RSU with sensing capability. Regarding the traffic simulation, there are several non-public simulators which can explicitly generate data tailored for collaborative perception such as scenes with occlusion, and sensor range limitations}~\cite{suo2021trafficsim,tan2021scenegen}. \hl{Yet in this work, we use open-source CARLA-SUMO co-simulation for traffic flow simulation and data recording. Vehicles are spawned in CARLA via SUMO to roam around in the town with random routes. Hundreds of vehicles are spawned in different towns ($Town03$, $Town04$ and $Town05$ that have crossroads as well as T-junctions in both the crowded downtown and suburb highway). We record several log files in different towns. Afterwards, at different junctions, we read out 100 scenes from the log files. Each scene has a 20-second duration, and we choose $M (M=2,3,4,5)$ vehicles as well as one RSU in each scene as intelligent agents to share information. Example scenarios are shown in} Fig.~\ref{fig:scene} (a).

\subsection{Downstream tasks}
Our dataset not only supports single-agent perception tasks such as 3D object detection, tracking, image-/point-cloud-based semantic segmentation, depth estimation, but also enables collaborative perception such as collaborative 3D object detection, tracking, and collaborative BEV semantic segmentation in urban driving scenes. We provide a benchmark for collaborative perception algorithms.

\subsection{Dataset statistics}
\textbf{Annotation statistics.} 
We provide statistics on the annotations and objects to highlight the inclusiveness and diversity of our dataset. 
In Fig.~\ref{fig:scene} (b) we analyze the size of the cars' bounding boxes within a 70m range from ego vehicles in each frame. The great variation of car sizes indicates that our scenes contain a diverse set of car makes and models that well includes most of the common real-world vehicles. 
Figure~\ref{fig:scene} (c) shows the annotation count in each frame for vehicles within 0-30m, 30-50m, and 50-70m ranges from each ego vehicle. It suggests that our dataset features both crowded scenes (up to 100 annotations within 50-70m from the ego vehicle) and less crowded driving scenarios (as low as 10 annotations within 30m from the ego vehicle). Figure~\ref{fig:scene} (d) contains statistics on the number of LiDAR points per annotation for single-agent and multi-agent scenarios respectively. The number of total LiDAR points of each object annotation increases when there are more than one agents observing the same object. Specifically, for a single agent, there are 183.83 points in each annotation on average, but the number goes up to 875.59 points per annotation for multiple agents. 

\textbf{Scene features.}
We analyze the distance between each two ego vehicles for every frame, as shown in Fig.~\ref{fig:scene} (e). An overwhelming percentage of the ego vehicles are presented within 20-30 meters from each other, suggesting they are geographically closely connected. The speed of cars within 70m from ego vehicles are shown in Fig.~\ref{fig:scene} (f). Given the fact that our scenes are selected near intersections, we notice that a major fraction of vehicles are slower than 10 km/h. However, the maximum speed is as high as 90+ km/h. 
Figure~\ref{fig:scene} (g) shows the percentage of annotations observed by a certain number of ego vehicles, up to 5. Over 60\% of the annotations are observed by at least two ego vehicles.

\section{Collaborative Perception Benchmark}\label{sec:benchmark}

We benchmark three crucial perception tasks in autonomous driving within the collaboration setting: detection, tracking, and semantic segmentation. The three tasks have been extensively studied since they generate essential perception knowledge for autonomous vehicles to make safer decisions. For performance evaluation, we follow the same evaluation protocol of the three tasks in the single-agent scenario, except that we utilize the information shared by other agents while the single-agent perception do not have access to such information. We report the results in two scenarios: (1) V2V only, and (2) V2X (V2V + V2I). 

\textbf{Dataset format and split.} Our V2X-Sim dataset follows the same storage format of nuScenes~\cite{Caesar2020nuScenesAM}, an authoritative multi-modality single-agent autonomous driving dataset. nuScenes collected real-world single-agent driving data to promote the single-agent autonomous driving research; while we simulate multi-agent scenarios to facilitate the next-generation V2X-aided collaborative perception technology. Each scene in our dataset contains a 20 second traffic flow at a certain intersection of three CARLA towns, and the multi-modality multi-agent sensory streams are recorded at 5Hz, meaning each scene is composed of 100 frames. We generate 100 scenes with a total of 10,000 frames, and in each scene, 2-5 vehicles are selected as the collaboration agents. We use 8,000/1,000/1,000 frames for training/validation/testing respectively, \hl{and we ensure that there is no overlap in terms of the intersections across training/validation/testing set.} Each frame has data sampled from multiple agents (vehicles and RSU). \hl{There are 37,200 samples in the training set, 5,000 samples in the validation set, and 5,000 samples in the test set. The split is shared across tasks.}

\textbf{Implementation details.} Bird's-eye-view (BEV) is a widely used and powerful representation in autonomous driving because it can describe the surrounding objects and overall context via a compact top-down 2D map~\cite{Wu2020MotionNetJP}. Therefore, BEV-based representation is adopted in all three tasks: we use a 3D voxel grid to represent the 3D world, employ binary representation, and assign each voxel a positive label if the voxel contains point cloud data. Since the generated 3D voxel grid can be considered as a pseudo-image whose height dimension is the channel dimension, we can perform the efficient 2D convolution instead of the heavy 3D convolution. Specifically, we crop the points located in the region of $[-32, 32] \times [-32, 32] \times[-3, 2]$ meters for vehicles defined in the ego-vehicle Euclidean coordinate and $[-32, 32] \times [-32, 32] \times[-8, -3]$ meters for RSU. The width and length of each voxel are 0.25 meter, and the height is 0.4 meter, meaning the generated BEV-based pseudo-image has a dimension of $256\times256\times13$ ($W\times L\times H$). Note that the models in all the tasks consume the 3D BEV map and generate perception results in 2D BEV. 

\begin{figure*}[t]
    \begin{minipage}{.59\linewidth}
        \centering
        \includegraphics[width=1\textwidth]{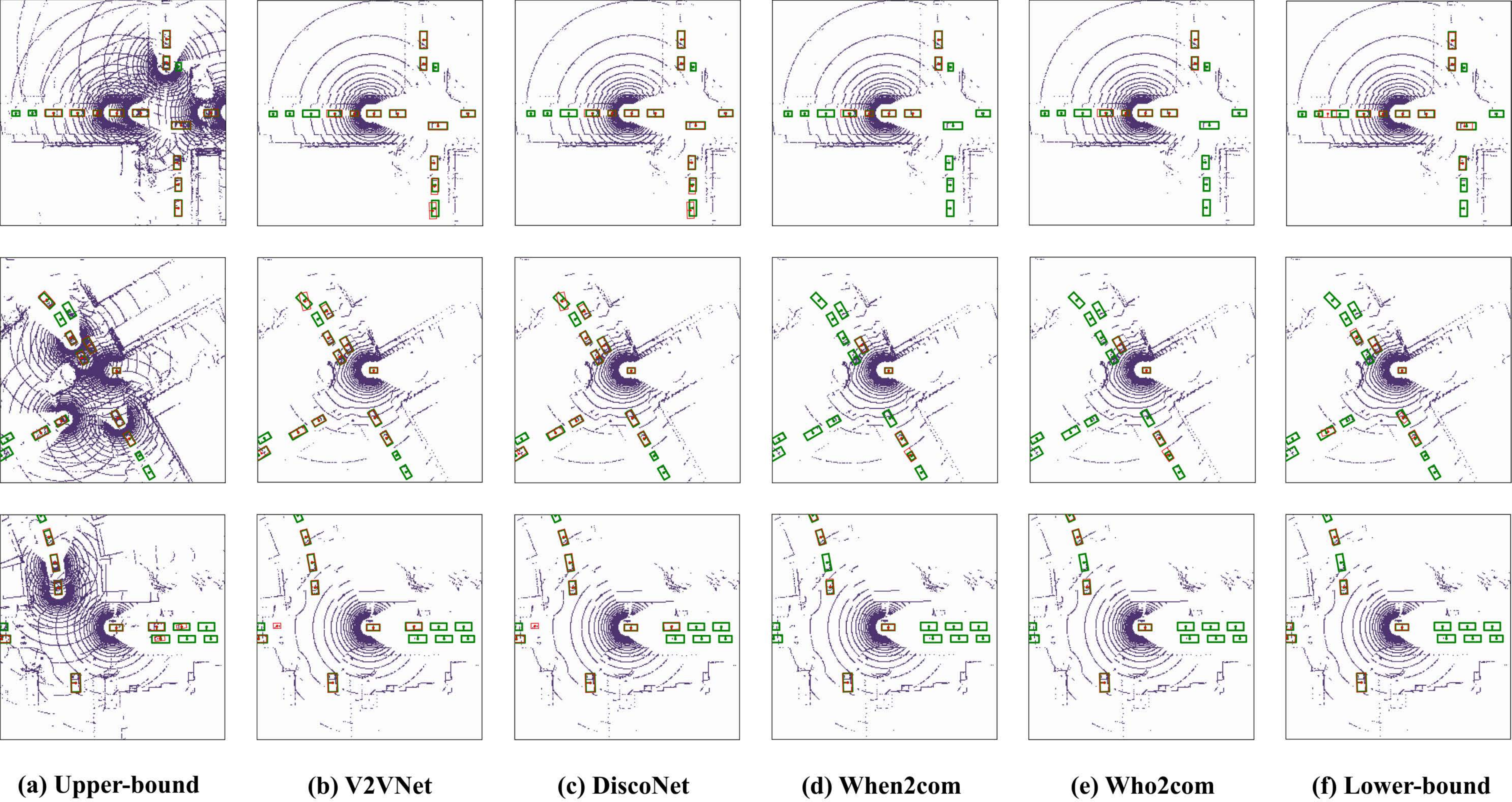}
        \captionsetup{font={scriptsize}}
        \caption{\hl{Visualizations of BEV detection on V2X-Sim.} \textcolor{red}{Red} and \textcolor{green}{green} boxes are the predictions and ground-truths respectively.}
        \label{fig:det-baseline}
    \end{minipage}~~
    \centering
    \begin{minipage}{.38\linewidth}
    \setlength{\tabcolsep}{0.1mm}{
    \captionsetup{font={scriptsize, sc, stretch=1.3}, justification=centering, labelsep=newline}
    \captionof{table}{\hl{Quantitative results of collaborative BEV detection on V2X-Sim. RSU denotes the Road-Side Unit. AP denotes the Average Precision. $\Delta$ is the absolute gain in AP introduced by RSU.}}
    \label{table:collaborative-bev-detection}
        \scriptsize
        \setlength{\extrarowheight}{0.3mm}{
        \begin{tabular}{cccccccc}
        \toprule
        \multicolumn{2}{c}{\multirow{2}{*}{\textbf{Method}}} & \multicolumn{3}{c}{\textbf{AP@IoU=0.5}} & \multicolumn{3}{c}{\textbf{AP@IoU=0.7}} \\
        \multicolumn{2}{c}{} & w/o RSU & w/ RSU & $\Delta$ & w/o RSU & w/ RSU & $\Delta$  \\ \midrule
        \multicolumn{2}{c}{Lower-bound} & 49.90 & 46.96 & $\downarrow$ 2.94 & 44.21 & 42.33 & $\downarrow$ 1.88 \\
        \multicolumn{2}{c}{Co-lower-bound} & 43.99 & 42.98 & $\downarrow$ 1.01 & 39.10 & 38.26 & $\downarrow$ 0.84 \\
        \multicolumn{2}{c}{When2com~\cite{Liu2020When2comMP}} & 44.02 & 46.39 & $\uparrow$ 2.37 & 39.89 & 40.32 & $\uparrow$ 0.43 \\
        \multicolumn{2}{c}{When2com*~\cite{Liu2020When2comMP}} & 45.35 & 48.28 & $\uparrow$ 2.93 & 40.45 & 41.43 & $\uparrow$ 0.68 \\
        \multicolumn{2}{c}{Who2com~\cite{Liu2020Who2comCP}} & 44.02 & 46.39 & $\uparrow$ 2.37  & 39.89 & 40.32 & $\uparrow$ 0.43 \\
        \multicolumn{2}{c}{Who2com*~\cite{Liu2020Who2comCP}} & 45.35 & 48.28 & $\uparrow$ 2.93 & 40.45 & 41.13 & $\uparrow$ 0.68  \\
        \multicolumn{2}{c}{V2VNet~\cite{Wang2020V2VNetVC}} & \textbf{\textcolor{blue}{68.35}} & \textbf{\textcolor{blue}{72.08}} & $\uparrow$ 3.73 & \textbf{\textcolor{blue}{62.83}} & \textbf{\textcolor{blue}{65.85}} & $\uparrow$ 3.02  \\
        \multicolumn{2}{c}{DiscoNet~\cite{Li_2021_NeurIPS}} & \textbf{\textcolor{green}{69.03}} & \textbf{\textcolor{green}{72.87}} & $\uparrow$ 3.84 & \textbf{\textcolor{green}{63.44}} & \textbf{\textcolor{green}{66.40}} & $\uparrow$ 4.81 \\
        \multicolumn{2}{c}{Upper-bound} & \textbf{\textcolor{red}{70.43}} & \textbf{\textcolor{red}{77.08}} & $\uparrow$ 6.65 & \textbf{\textcolor{red}{67.04}} & \textbf{\textcolor{red}{72.57}} & $\uparrow$ 5.53  \\
        \bottomrule
        \end{tabular}}}
    \end{minipage}~~
\vspace{-2mm}
\end{figure*}

\textbf{Benchmark models.} We aim to benchmark collaborative perception strategies rather than the well-studied individual perception methods. We consider \hl{early/intermediate/late/no collaboration} models for the benchmark. The intermediate models, including DiscoNet~\cite{Li_2021_NeurIPS}, V2VNet~\cite{Wang2020V2VNetVC}, When2com~\cite{Liu2020When2comMP}, and Who2com~\cite{Liu2020Who2comCP}, are based on the communication of the intermediate features of the neural network. The methods in our benchmark are as follows:
\begin{itemize}
    \item \textbf{Lower-bound}: The single-agent perception model without collaboration which processes a single-view point cloud is considered as the lower-bound. 
    
    \item \hl{\textbf{Co-lower-bound}: Collaborative lower-bound fuses the output from different single-agent perception models. }
    
    \item \textbf{Upper-bound}: The early collaboration model which transmits raw point cloud data is the upper-bound.
     \item \textbf{DiscoNet~\cite{Li_2021_NeurIPS}}: DiscoNet uses a directed collaboration graph with matrix-valued edge weight to adaptively highlight the informative spatial regions and reject the noisy regions of the messages sent by the partners. After adaptive message fusion, the updated features will be transmitted to the output head for perception. 
    \item \textbf{V2VNet~\cite{Wang2020V2VNetVC}}: V2VNet uses a pose-aware graph neural network to propagate agents’ information, and employs a convolutional gated recurrent unit based module to aggregate other agents’ information. After several rounds of neural message passing, the updated features are fed into the output head to generate perception results.
    \item \textbf{When2com~\cite{Liu2020When2comMP}}: When2com employs attention-based mechanism for communication group construction: the partners with satisfactory correlation scores will be selected as the collaborators. After the attention-score-based weighted fusion, the updated features will be fed into the output head for perceptions. The model with pose information is marked by $*$.
    \item \textbf{Who2com~\cite{Liu2020Who2comCP}}: Who2com shares a similar idea with When2com, yet it uses handshake mechanism to select the collaborator: the partner with the highest score will be selected as the collaborator. The model with pose information is marked by $*$.
\end{itemize}

We implement a 3D perception pipeline that can be integrated with all of the communication methods mentioned above. Since the source codes of V2VNet is not publicly available, we re-implement the V2VNet in PyTorch according to its pseudo-code. For when2com/who2com, we borrow their communication modules from its official code. All of the intermediate collaboration modules use the same architecture and conduct collaboration at the same intermediate feature layer. Moreover, all of the methods are trained with the same setting to ensure that the performance gain comes from the collaboration instead of irrelevant techniques. 

\begin{table*}[t]
\scriptsize
\centering
\captionsetup{font=sc}
\captionsetup{font={scriptsize, sc, stretch=1.3}, justification=centering, labelsep=newline}
\caption{\hl{Quantitative results of BEV tracking on V2X-Sim. \textbf{MOTA}: Multiple Object Tracking Accuracy. \textbf{MOTP}: Multiple Object Tracking Precision. \textbf{HOTA}: Higher Order
Tracking Accuracy. \textbf{DetA}: Detection Accuracy. \textbf{AssA}: Association Accuracy. \textbf{DetRe}: Detection Recall. \textbf{DetPr}: Detection Precision. \textbf{AssRe}: Association Recall. \textbf{AssPr}: Association Precision. \textbf{LocA}: Localization Accuracy. The number to the left of () denotes the performance in V2V solely. The number in () represents the performance gain by adding V2I.}}
\label{table:tracking-baseline}
\setlength{\tabcolsep}{1mm}{
\begin{tabular}{@{}cccccccccccc@{}}
\toprule
\textbf{Method} & \textbf{MOTA} & \textbf{MOTP} & \textbf{HOTA} & \textbf{DetA} & \textbf{AssA} & \textbf{DetRe} & \textbf{DetPr} & \textbf{AssRe} & \textbf{AssPr} & \textbf{LocA} \\ \midrule
Lower-bound & 35.72 ($\downarrow$3.87) & 84.16 ($\downarrow$0.74) & 34.27 ($\downarrow$1.68) & 33.64 ($\downarrow$3.24) & 36.18 ($\downarrow$0.06) & 35.07 ($\downarrow$3.54) & 82.49 ($\uparrow$0.96) & 46.70 ($\uparrow$0.23) & \textbf{\textcolor{blue}{58.72}} ($\uparrow$0.10) & 86.43 ($\uparrow$0.38) \\
Co-lower-bound & 21.53 ($\uparrow$0.58) & \textbf{\textcolor{blue}{85.76}} ($\uparrow$0.15) & 39.16 ($\downarrow$0.71) & 41.14 ($\downarrow$0.93) & 38.18 ($\downarrow$0.62) & \textbf{\textcolor{red}{59.54}} ($\downarrow$2.52) & 54.68 ($\uparrow$0.79) & \textbf{\textcolor{red}{50.92}} ($\downarrow$0.65) & 55.78 ($\uparrow$0.84) & \textbf{\textcolor{blue}{87.64}} ($\uparrow$0.38)\\
When2com~\cite{Liu2020When2comMP} & 29.48 ($\uparrow$2.45) & \textbf{\textcolor{green}{86.10}} ($\downarrow$2.79) & 30.94 ($\uparrow$1.01) & 27.90 ($\uparrow$2.04) & 35.33 ($\uparrow$0.06) & 28.67 ($\uparrow$2.58) & \textbf{\textcolor{blue}{86.11}} ($\downarrow$4.81) & 46.30 ($\downarrow$0.15) & \textbf{\textcolor{red}{59.20}} ($\downarrow$0.36) & \textbf{\textcolor{green}{87.98}} ($\downarrow$1.98)  \\
When2com$^*$~\cite{Liu2020When2comMP} & 30.17 ($\uparrow$1.43) & 84.95 ($\downarrow$1.44) & 31.34 ($\uparrow$0.43) & 29.11 ($\uparrow$1.05) & 35.42 ($\uparrow$0.21) & 30.28 ($\uparrow$1.32) & 83.81 ($\uparrow$0.29) & 46.65 ($\downarrow$0.29) & 58.61 ($\uparrow$0.18) & 86.14 ($\uparrow$0.17) \\
Who2com~\cite{Liu2020Who2comCP} & 29.48 ($\uparrow$2.46) & \textbf{\textcolor{green}{86.10}} ($\downarrow$2.79) & 30.94 ($\uparrow$1.01) & 27.90 ($\uparrow$2.04) & 35.33 ($\uparrow$0.06) & 28.67 ($\uparrow$2.58) & \textbf{\textcolor{blue}{86.11}} ($\downarrow$4.81) & 46.30 ($\downarrow$0.15) & \textbf{\textcolor{red}{59.20}} ($\downarrow$0.36) & \textbf{\textcolor{green}{87.98}} ($\downarrow$1.98)  \\
Who2com$^*$~\cite{Liu2020Who2comCP} & 30.17 ($\uparrow$1.43) & 84.95 ($\downarrow$1.44) & 31.34 ($\uparrow$0.43) & 29.11 ($\uparrow$1.06) & 35.42 ($\uparrow$0.21) & 30.28 ($\uparrow$1.33) & 83.81 ($\uparrow$0.29) & 46.65 ($\downarrow$0.29) & 58.61 ($\uparrow$0.18) & 86.14 ($\uparrow$0.17) \\
V2VNet~\cite{Wang2020V2VNetVC} & \textbf{\textcolor{blue}{55.29}} ($\uparrow$2.29) & 85.21 ($\downarrow$0.53) & \textbf{\textcolor{blue}{43.68}} ($\uparrow$0.91) & \textbf{\textcolor{blue}{50.71}} ($\uparrow$1.93) & \textbf{\textcolor{blue}{38.76}} ($\uparrow$0.24) & 53.40 ($\uparrow$2.51) & 84.45 ($\downarrow$1.07) & 50.22 ($\uparrow$0.53) & 58.50 ($\downarrow$0.07) & 87.22 ($\uparrow$0.38) \\
DiscoNet~\cite{Li_2021_NeurIPS} & \textbf{\textcolor{green}{56.69}} ($\uparrow$2.26) & \textbf{\textcolor{red}{86.23}} ($\downarrow$0.41) & \textbf{\textcolor{green}{44.76}} ($\uparrow$1.09) & \textbf{\textcolor{green}{52.41}} ($\uparrow$2.18) & \textbf{\textcolor{red}{39.25}} ($\uparrow$1.11) & \textbf{\textcolor{blue}{54.87}} ($\uparrow$2.58) & \textbf{\textcolor{green}{86.29}} ($\downarrow$0.95) & \textbf{\textcolor{green}{50.86}} ($\uparrow$1.02) & \textbf{\textcolor{green}{58.94}} ($\downarrow$0.15) & \textbf{\textcolor{red}{88.07}} ($\uparrow$0.34)\\
Upper-bound & \textbf{\textcolor{red}{58.00}} ($\uparrow$3.92) & 85.61 ($\uparrow$0.25) & \textbf{\textcolor{red}{44.83}} ($\uparrow$4.24) & \textbf{\textcolor{red}{52.94}} ($\uparrow$4.24) & \textbf{\textcolor{green}{38.95}} ($\downarrow$0.75) & \textbf{\textcolor{green}{55.07}} ($\uparrow$4.68) & \textbf{\textcolor{red}{86.54}} ($\downarrow$0.30) & \textbf{\textcolor{blue}{50.35}} ($\downarrow$0.86) & 58.71 ($\uparrow$0.15) & 87.48 ($\uparrow$0.06) \\
\bottomrule
\vspace{-8mm}
\end{tabular}}
\end{table*}

\subsection{Collaborative object detection in BEV}
\textbf{Problem definition.} As the most crucial perception task in autonomous driving, 3D object detection aims to recognize and localize the objects in 3D scenes given a single frame, with the following tracking, prediction and planning modules all heavily relying on the detections. The models consume voxelized point cloud and output BEV bounding boxes.

\textbf{Backbone and evaluation.} We use a classic anchor-based detector composed of a convolutional encoder, a convolutional decoder, and an output header for classification and regression~\cite{Luo2018FastAF}. Regarding the loss function, we use the binary cross-entropy loss to supervise the box classification and the smooth $L_1$ loss to supervise the box regression, following~\cite{Luo2018FastAF}. We employ the generic BEV detection evaluation metric: Average Precision (AP) at Intersection-over-Union (IoU) thresholds of 0.5 and 0.7. We target the vehicle detection and report the results on the test set.

\textbf{Quantitative results.} Table~\ref{table:collaborative-bev-detection} demonstrates the quantitative comparisons on AP (@IoU = 0.5/0.7). We find that: (1) the upper-bound performs best amongst all methods, and it improves lower-bound significantly by 41.1$\%$ and 51.6$\%$ in terms of AP@0.5 and AP@0.7 in the scenario of V2V only, validating the necessity of collaboration; (2) V2V and V2I jointly can generally improve the perception over V2V only with more viewpoints, e.g., adding V2I can bring an improvement of 9.4$\%$ for upper-bound, and 5.5$\%$ for V2VNet in terms of AP@0.5; (3) among the intermediate models, DiscoNet achieves the best performance via the well-designed distilled collaboration graph, V2VNet achieves the second best performance via the powerful neural message passing, and When2com/Who2com only achieve comparable performance with lower-bound since the attention-mechanism is not suitable in point-cloud-based collaborative perception: the agents usually need complementary information rather than a highly similar one; \hl{ (4) the late collaboration model (co-lower-bound) hurts the detection performance because of introducing extra false positives from other vehicles.}

\textbf{Qualitative results.} The qualitative results are shown in Fig.~\ref{fig:det-baseline}. We see that the collaboration can fundamentally mitigate the problems of long-range perception and occlusion. 

\begin{figure*}[t]
\begin{center}
\hsize=\textwidth
\includegraphics[width=0.87\textwidth]{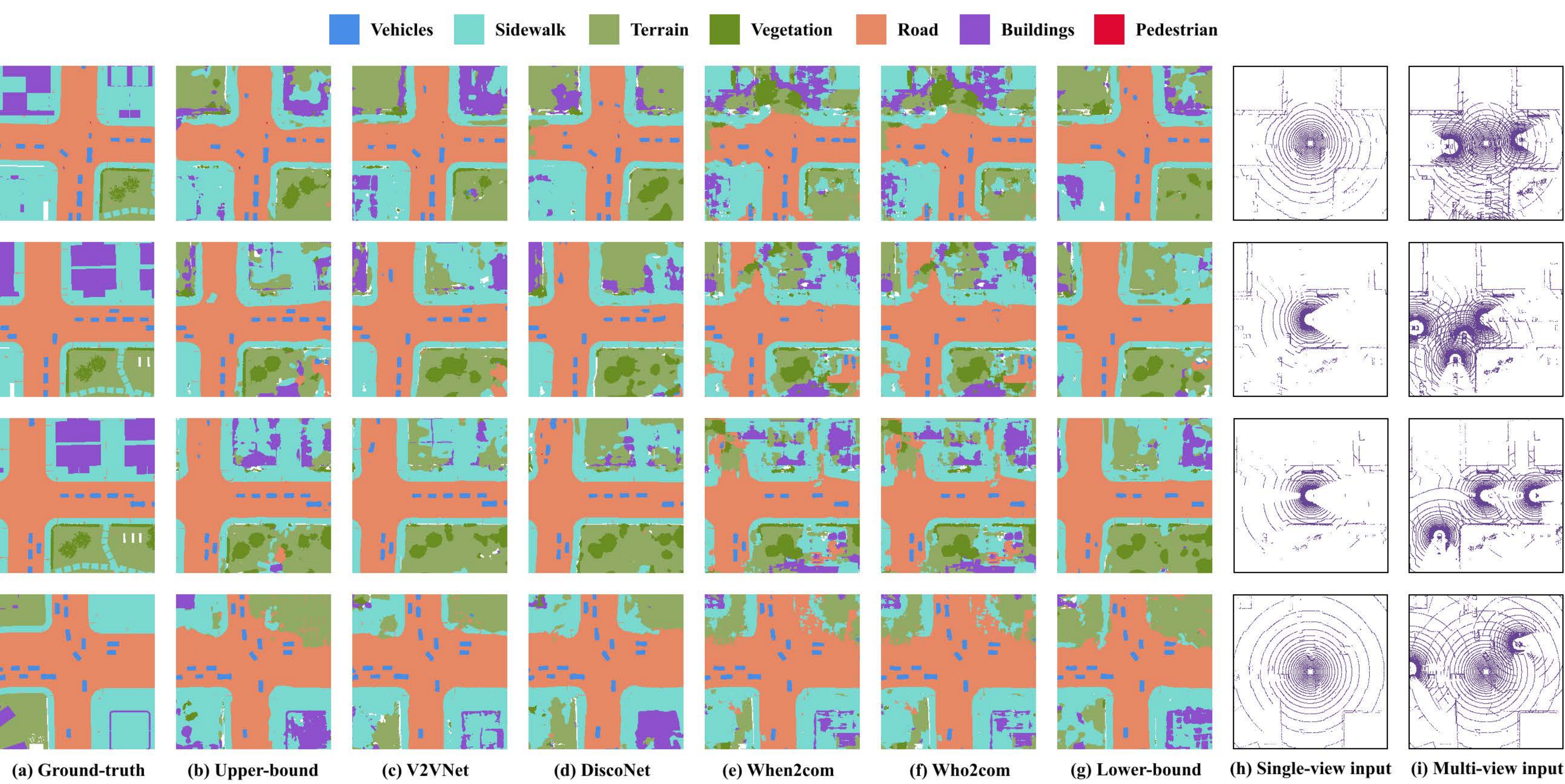}
\captionsetup{font={scriptsize}}
\caption{\hl{Visualizations of collaborative BEV semantic segmentation.}}
\label{fig:seg-baseline}
\end{center}
\vspace{-5mm}
\end{figure*}

\begin{table*}[t]
\scriptsize
\centering
\captionsetup{font={scriptsize, sc, stretch=1.3}, justification=centering, labelsep=newline}
\caption{\hl{Quantitative results of BEV segmentation on V2X-Sim. The number to the left of () denotes the performance in V2V solely. The number in () represents the performance gain by adding V2I.}}
\label{table:segmentation-baseline}
\setlength{\tabcolsep}{1.8mm}{
\begin{tabular}{@{}ccccccccc@{}}
\toprule
\textbf{Method} & \textbf{Vehicle} & \textbf{Sidewalk} & \textbf{Terrain} & \textbf{Road} & \textbf{Building} & \textbf{Pedestrian} & \textbf{Vegetation} & \textbf{mIoU} \\ \midrule
Lower-bound & 45.93 ($\uparrow$2.22) & 42.39 ($\downarrow$2.75) & 47.03 ($\uparrow$0.20) & 65.76 ($\downarrow$1.27) & 25.38 ($\downarrow$1.89) & 20.59 ($\downarrow$3.09) & 35.83 ($\uparrow$0.66) & 36.64 ($\downarrow$0.87)\\
Co-lower-bound & 47.67 ($\uparrow$2.43) & \textbf{\textcolor{green}{48.79}} ($\downarrow$1.41) & \textbf{\textcolor{red}{50.92}} ($\uparrow$0.85) & \textbf{\textcolor{green}{70.00}} ($\downarrow$0.65) & 25.26 ($\uparrow$0.17) & 10.78 ($\downarrow$1.77) & 39.46 ($\uparrow$2.69) & 38.38 ($\uparrow$0.46)\\
When2com~\cite{Liu2020When2comMP} & 48.43 ($\uparrow$0.03) & 33.06 ($\uparrow$1.38) & 36.89 ($\uparrow$1.76) & 57.74 ($\uparrow$1.56) & \textbf{\textcolor{red}{29.20}} ($\uparrow$1.18) & 20.37 ($\uparrow$0.57) & 39.17 ($\downarrow$0.01) & 34.49 ($\uparrow$0.88)\\
When2com$*$~\cite{Liu2020When2comMP} & 47.74 ($\uparrow$1.23) & 33.60 ($\downarrow$0.40) & 35.81 ($\uparrow$1.05) & 56.75 ($\uparrow$0.48) & 26.11 ($\downarrow$0.92) & 19.16 ($\uparrow$0.04) & 39.64 ($\downarrow$2.55) & 33.81 ($\downarrow$0.47)\\
Who2com~\cite{Liu2020Who2comCP} & 48.40 ($\uparrow$0.06) & 32.76 ($\uparrow$1.68) & 36.04 ($\uparrow$2.61) & 57.51 ($\uparrow$1.79) & \textbf{\textcolor{green}{29.17}} ($\uparrow$1.21) & 20.36 ($\uparrow$0.58) & 39.08 ($\uparrow$0.08) & 34.31 ($\uparrow$1.06) \\
Who2com$*$~\cite{Liu2020Who2comCP} & 47.74 ($\uparrow$1.23) & 33.60 ($\downarrow$0.40) & 35.81 ($\uparrow$1.05) & 56.75 ($\uparrow$0.48) & 26.11 ($\downarrow$0.92) & 19.16 ($\uparrow$0.04) & 39.64 ($\downarrow$2.55) & 33.81 ($\downarrow$0.47) \\
V2VNet~\cite{Wang2020V2VNetVC} & \textbf{\textcolor{green}{58.42}} ($\uparrow$3.09) & \textbf{\textcolor{red}{48.33}} ($\downarrow$3.87) & 48.51 ($\downarrow$1.59) & \textbf{\textcolor{red}{70.02}} ($\uparrow$0.46) & 28.58 ($\uparrow$5.18) & \textbf{\textcolor{blue}{21.99}} ($\uparrow$0.57) & \textbf{\textcolor{blue}{41.42}} ($\uparrow$0.35) & \textbf{\textcolor{green}{41.11}} ($\uparrow$0.74) \\ 
DiscoNet~\cite{Li_2021_NeurIPS} & \textbf{\textcolor{blue}{56.66}} ($\uparrow$1.19) & \textbf{\textcolor{blue}{46.98}} ($\downarrow$1.74) & \textbf{\textcolor{green}{50.22}} ($\downarrow$1.05) & \textbf{\textcolor{blue}{68.62}} ($\downarrow$0.25) & 27.36 ($\uparrow$5.58) & \textbf{\textcolor{green}{22.02}} ($\uparrow$0.82) & \textbf{\textcolor{green}{42.50}} ($\uparrow$0.95) & \textbf{\textcolor{blue}{40.84}} ($\uparrow$0.53) \\ 
Upper-bound & \textbf{\textcolor{red}{64.09}} ($\uparrow$5.34) & 41.34 ($\uparrow$2.42) & \textbf{\textcolor{blue}{48.20}} ($\uparrow$0.74) & 67.05 ($\uparrow$2.04) & \textbf{\textcolor{blue}{29.07}} ($\uparrow$0.74) & \textbf{\textcolor{red}{31.54}} ($\uparrow$3.15) & \textbf{\textcolor{red}{45.04}} ($\uparrow$0.70) & \textbf{\textcolor{red}{42.29}} ($\uparrow$1.98)\\
\bottomrule
\end{tabular}}
\vspace{-5mm}
\end{table*}

\subsection{Collaborative multi-object tracking in BEV}
\textbf{Problem definition.}
Different from detection, multi-object tracking requires the generation of temporally consistent perception results. Multi-object tracking in BEV is to use bounding boxes, object categories, and object identities to track different objects within a temporal sequence. 

\textbf{Evaluation metrics.} 
We mainly utilize HOTA (Higher Order Tracking Accuracy)~\cite{Luiten_2020} to evaluate our BEV tracking performance. HOTA can evaluate detection, association, and
localization performance via a single unified metric. In addition, the classic multi-object tracking accuracy (MOTA) and multi-object tracking precision (MOTP)~\cite{journals/ejivp/BernardinS08} are also employed. MOTA can measure detection errors and association errors. MOTP solely measures localization accuracy. 

\textbf{Baseline tracker.} 
We implement SORT~\cite{Bewley2016_sort} as our baseline tracker. Given the detection results, SORT will combine the Kalman Filter and Hungarian algorithm to achieve an accurate and efficient tracking performance.

\textbf{Quantitative results.} Quantitative comparisons of BEV tracking are shown in Table~\ref{table:tracking-baseline}. Similar to BEV detection, upper-bound achieves the best performance in terms of MOTA and HOTA. Meanwhile, adding V2I can improve MOTA largely yet cannot help too much in MOTP. Co-lower-bound shows good performance in localization accuracy (MOTP). A more advanced tracker is required to exploit the collaboration for filling the performance gap.

\subsection{Collaborative semantic segmentation in BEV}

\textbf{Problem definition.} 
We aim to conduct semantic segmentation in BEV using only geometry point cloud. In the collaborative perception scenarios, there are measurements collected by multiple agents with distinct viewpoints. Therefore, there are more information in the scene, facilitating the semantic scene understanding.

\textbf{Baseline segmentation method and evaluation metrics.} 
We follow the backbone architecture as well as the loss function of U-Net~\cite{Ronneberger2015} in our baseline method. The input is a BEV-based voxelized point cloud, and the output is BEV semantic segmentation. We label and predict seven categories as listed in Table~\ref{table:segmentation-baseline}, while the remaining is unlabeled. In our benchmark, we evaluate the segmentation performance using mean Intersection-over-Union (mIoU).

\textbf{Quantitative results.} As illustrated in Table \ref{table:segmentation-baseline}, we find that: (1) V2VNet, DiscoNet, and upper-bound achieve comparable performance in terms of terrain and road categories; \hl{(2) the attention-based methods (i.e., when2com, who2com) performs worse because the attention-based mechanisms try to find the collaboration partners with high correlation scores. Whereas, in 3D perception, the collaborators with complementary information should be prioritized during the collaboration. Such nonalignment can make it quite hard for the attention model to learn;} (3) there is a large gap between lower-bound and upper-bound regarding the two safety-critical categories: vehicle ($45.93\%$ v.s. $64.09\%$) and pedestrian ($20.59\%$ v.s. $31.54\%$), proving the values of collaboration; \hl{(4) employing V2V and V2I jointly can generally enhance the vehicle segmentation over using V2V solely; (5) co-lower-bound  performs better than the lower-bound.}

\textbf{Qualitative results.}
Figure \ref{fig:seg-baseline} shows the semantics segmentation results. The results of upper-bound restore the semantic information with rich point cloud data. Intermediate-based collaboration strategies V2VNet and DiscoNet can achieve satisfactory performance yet When2com and Who2com hurt the performance compared to the lower-bound.

\begin{figure*}[t]
\begin{center}
\hsize=\textwidth
\includegraphics[width=0.85\textwidth]{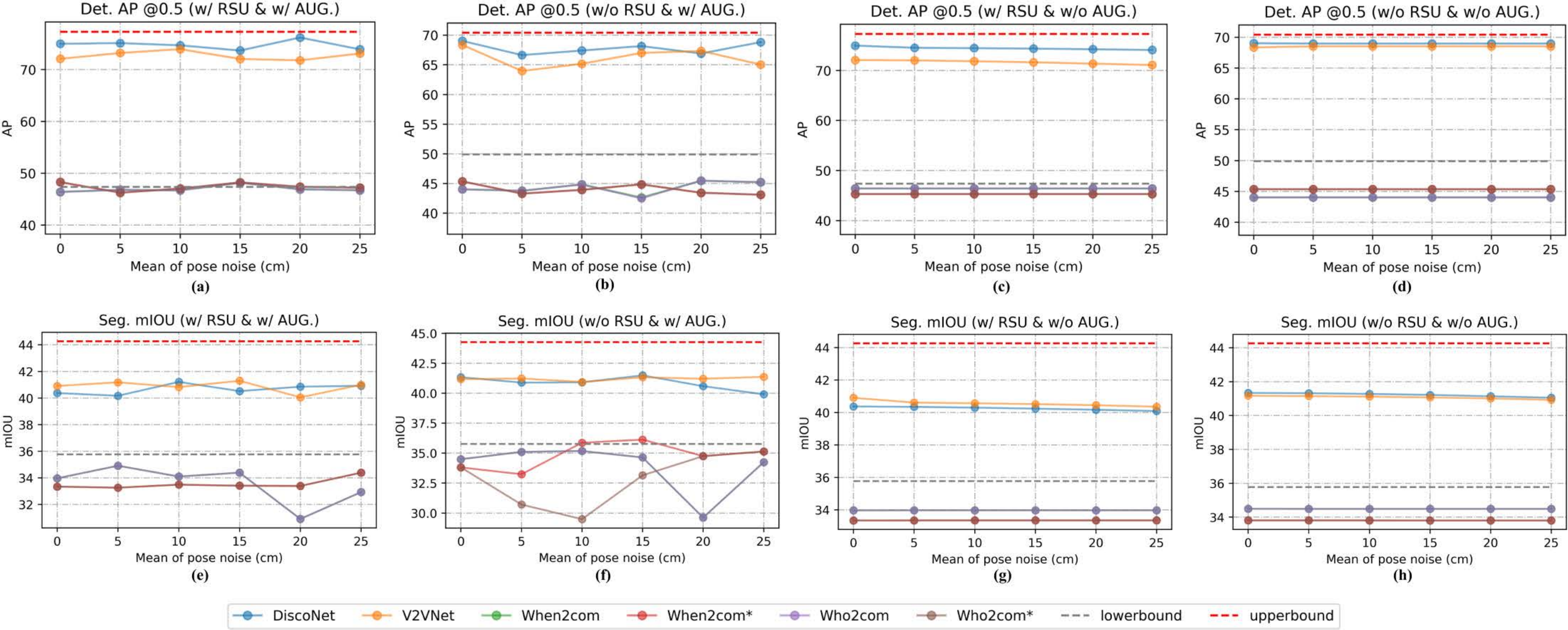}
\captionsetup{font={scriptsize}}
\caption{\hl{Experimental results of a robustness test conducted under various magnitudes of pose noise. AUG. means augmentation which adds pose noise during training.}}
\label{fig:pose}
\end{center}
\end{figure*}

\begin{figure*}[t]
\begin{center}
\hsize=\textwidth
\includegraphics[width=0.85\textwidth]{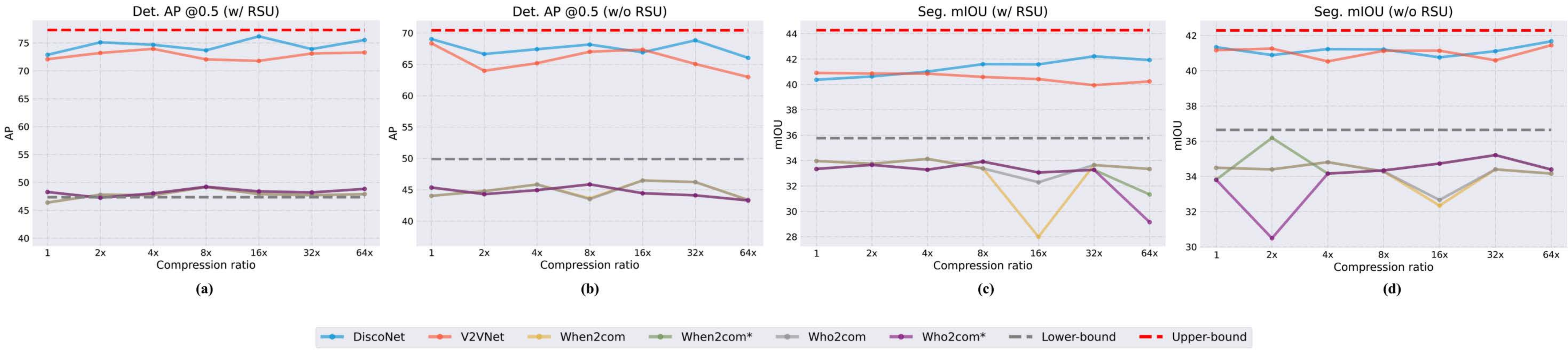}
\captionsetup{font={scriptsize}}
\caption{\hl{Experimental results of a bandwidth test conducted under various compression ratios. }}
\label{fig:compression}
\end{center}
\vspace{-5mm}
\end{figure*}

\subsection{Discussions on pose noise and compression ratio}
\hl{We further examine the robustness of different intermediate models against realistic pose noise (Gaussian noise with a mean of $0.05m-0.25m$ and a standard deviation of $0.02cm$), as shown in Fig.}~\ref{fig:pose}. \hl{We can see that the models perform comparably with or without pose noise in the training phase, and all the intermediate methods have shown stable performance against the pose noise. The reason is that the intermediate feature map has a relatively low spatial resolution (each grid in the feature map denotes a coverage of $2m\times2m$), thus is not vulnerable to noisy pose. Meanwhile, we employ an $1\times1$ autoencoder to further compress the feature channel of the transmitted feature map. We test the models with different compression ratios, and we found that the jointly learned $1\times1$ autoencoder can even improve the performance a little bit, and most intermediate models achieve comparable performance at different levels of compression, as shown in Fig.}~\ref{fig:compression}.

\section{Conclusion}\label{sec:conclusion}
We propose V2X-Sim dataset based on CARLA-SUMO co-simulation, in order to enable multi-agent collaborative perception research in autonomous driving. Our dataset provides multi-agent multi-modality sensor streams captured by the vehicles and road-side unit (RSU) in realistic traffic flows. Diverse annotations are provided to support a variety of 3D perception tasks. In addition, we benchmark several state-of-the-art collaborative perception methods in collaborative BEV detection, tracking, and semantic segmentation tasks. Future works include the simulation of latency issues as well as the development of novel evaluation metrics in collaborative perception tasks. We believe our work can inspire many relevant research areas including but not limited to autonomous driving, computer vision, multi-robot system, communication engineering, and machine learning. 

\textbf{Acknowledgment.} The authors would like to thank anonymous reviewers for their helpful suggestions, and NYU high performance computing (HPC) for the support.





\bibliographystyle{IEEEtran}  
\normalem
\bibliography{IEEEabrv,ref}

\begin{thebibliography}{10}
\providecommand{\url}[1]{#1}
\csname url@rmstyle\endcsname
\providecommand{\newblock}{\relax}
\providecommand{\bibinfo}[2]{#2}
\providecommand\BIBentrySTDinterwordspacing{\spaceskip=0pt\relax}
\providecommand\BIBentryALTinterwordstretchfactor{4}
\providecommand\BIBentryALTinterwordspacing{\spaceskip=\fontdimen2\font plus
\BIBentryALTinterwordstretchfactor\fontdimen3\font minus
  \fontdimen4\font\relax}
\providecommand\BIBforeignlanguage[2]{{%
\expandafter\ifx\csname l@#1\endcsname\relax
\typeout{** WARNING: IEEEtran.bst: No hyphenation pattern has been}%
\typeout{** loaded for the language `#1'. Using the pattern for}%
\typeout{** the default language instead.}%
\else
\language=\csname l@#1\endcsname
\fi
#2}}

\bibitem{Caesar2020nuScenesAM}
H.~Caesar, V.~Bankiti, A.~Lang, S.~Vora, V.~E. Liong, Q.~Xu, A.~Krishnan,
  Y.~Pan, G.~Baldan, and O.~Beijbom, ``nuscenes: A multimodal dataset for
  autonomous driving,'' in \emph{Proc. IEEE Conf. Comput. Vis. Pattern
  Recognit.}, 2020, pp. 11\,618--11\,628.

\bibitem{Geiger2012AreWR}
A.~Geiger, P.~Lenz, and R.~Urtasun, ``Are we ready for autonomous driving? the
  kitti vision benchmark suite,'' in \emph{Proc. IEEE Conf. Comput. Vis.
  Pattern Recognit.}, 2012, pp. 3354--3361.

\bibitem{Sun2020ScalabilityIP}
P.~Sun, H.~Kretzschmar, X.~Dotiwalla, A.~Chouard, V.~Patnaik, P.~Tsui, J.~Guo,
  Y.~Zhou, Y.~Chai, B.~Caine, V.~Vasudevan, W.~Han, J.~Ngiam, H.~Zhao,
  A.~Timofeev, S.~Ettinger, M.~Krivokon, A.~Gao, A.~Joshi, Y.~Zhang, J.~Shlens,
  Z.-F. Chen, and D.~Anguelov, ``Scalability in perception for autonomous
  driving: Waymo open dataset,'' in \emph{Proc. IEEE Conf. Comput. Vis. Pattern
  Recognit.}, 2020, pp. 2443--2451.

\bibitem{arnold2019survey}
E.~Arnold, O.~Y. Al-Jarrah, M.~Dianati, S.~Fallah, D.~Oxtoby, and
  A.~Mouzakitis, ``A survey on 3d object detection methods for autonomous
  driving applications,'' \emph{IEEE Transactions on Intelligent Transportation
  Systems}, vol.~20, no.~10, pp. 3782--3795, 2019.

\bibitem{marvasti2021deep}
S.~M. Marvasti-Zadeh, L.~Cheng, H.~Ghanei-Yakhdan, and S.~Kasaei, ``Deep
  learning for visual tracking: A comprehensive survey,'' \emph{IEEE
  Transactions on Intelligent Transportation Systems}, 2021.

\bibitem{Minaee2021ImageSU}
S.~Minaee, Y.~Boykov, F.~Porikli, A.~Plaza, N.~Kehtarnavaz, and D.~Terzopoulos,
  ``Image segmentation using deep learning: A survey,'' \emph{IEEE transactions
  on pattern analysis and machine intelligence}, vol.~PP, 2021.

\bibitem{machardy2018v2x}
Z.~MacHardy, A.~Khan, K.~Obana, and S.~Iwashina, ``V2x access technologies:
  Regulation, research, and remaining challenges,'' \emph{IEEE Communications
  Surveys \& Tutorials}, vol.~20, no.~3, pp. 1858--1877, 2018.

\bibitem{muhammad2018survey}
M.~Muhammad and G.~A. Safdar, ``Survey on existing authentication issues for
  cellular-assisted v2x communication,'' \emph{Vehicular Communications},
  vol.~12, pp. 50--65, 2018.

\bibitem{hasan2020securing}
M.~Hasan, S.~Mohan, T.~Shimizu, and H.~Lu, ``Securing vehicle-to-everything
  (v2x) communication platforms,'' \emph{IEEE Transactions on Intelligent
  Vehicles}, vol.~5, no.~4, pp. 693--713, 2020.

\bibitem{mannoni2019comparison}
V.~Mannoni, V.~Berg, S.~Sesia, and E.~Perraud, ``A comparison of the v2x
  communication systems: Its-g5 and c-v2x,'' in \emph{2019 IEEE 89th Vehicular
  Technology Conference (VTC2019-Spring)}.\hskip 1em plus 0.5em minus
  0.4em\relax IEEE, 2019, pp. 1--5.

\bibitem{Chen2019CooperCP}
Q.~Chen, S.~Tang, Q.~Yang, and S.~Fu, ``Cooper: Cooperative perception for
  connected autonomous vehicles based on 3d point clouds,'' in \emph{IEEE 39th
  International Conference on Distributed Computing Systems (ICDCS)}, 2019, pp.
  514--524.

\bibitem{Wang2020V2VNetVC}
T.-H. Wang, S.~Manivasagam, M.~Liang, B.~Yang, W.~Zeng, J.~Tu, and R.~Urtasun,
  ``V2vnet: Vehicle-to-vehicle communication for joint perception and
  prediction,'' in \emph{ECCV}, 2020.

\bibitem{Li_2021_NeurIPS}
Y.~Li, S.~Ren, P.~Wu, S.~Chen, C.~Feng, and W.~Zhang, ``Learning distilled
  collaboration graph for multi-agent perception,'' in \emph{NeurIPS}, 2021.

\bibitem{yuan2021comap}
Y.~Yuan and M.~Sester, ``Comap: A synthetic dataset for collective multi-agent
  perception of autonomous driving,'' \emph{The International Archives of
  Photogrammetry, Remote Sensing and Spatial Information Sciences}, vol.~43,
  pp. 255--263, 2021.

\bibitem{yuan2022keypoints}
Y.~Yuan, H.~Cheng, and M.~Sester, ``Keypoints-based deep feature fusion for
  cooperative vehicle detection of autonomous driving,'' \emph{IEEE Robotics
  and Automation Letters}, vol.~7, no.~2, pp. 3054--3061, 2022.

\bibitem{krajzewicz2012recent}
D.~Krajzewicz, J.~Erdmann, M.~Behrisch, and L.~Bieker, ``Recent development and
  applications of sumo-simulation of urban mobility,'' \emph{International
  journal on advances in systems and measurements}, vol.~5, no. 3\&4, 2012.

\bibitem{Dosovitskiy17}
A.~Dosovitskiy, G.~Ros, F.~Codevilla, A.~Lopez, and V.~Koltun, ``{CARLA}: {An}
  open urban driving simulator,'' in \emph{Proceedings of the 1st Annual
  Conference on Robot Learning}, 2017, pp. 1--16.

\bibitem{Liu2020When2comMP}
Y.-C. Liu, J.~Tian, N.~Glaser, and Z.~Kira, ``When2com: Multi-agent perception
  via communication graph grouping,'' in \emph{IEEE Conference on Computer
  Vision and Pattern Recognition (CVPR)}, 2020, pp. 4105--4114.

\bibitem{Liu2020Who2comCP}
Y.-C. Liu, J.~Tian, C.-Y. Ma, N.~Glaser, C.-W. Kuo, and Z.~Kira, ``Who2com:
  Collaborative perception via learnable handshake communication,'' in
  \emph{IEEE International Conference on Robotics and Automation (ICRA)}, 2020,
  pp. 6876--6883.

\bibitem{xu2021opv2v}
R.~Xu, H.~Xiang, X.~Xia, X.~Han, J.~Liu, and J.~Ma, ``Opv2v: An open benchmark
  dataset and fusion pipeline for perception with vehicle-to-vehicle
  communication,'' in \emph{ICRA}, 2022.

\bibitem{Xiao2018MultimediaFA}
Z.~Xiao, Z.~Mo, K.~Jiang, and D.~Yang, ``Multimedia fusion at semantic level in
  vehicle cooperactive perception,'' in \emph{IEEE International Conference on
  Multimedia \& Expo Workshops (ICMEW)}, 2018, pp. 1--6.

\bibitem{Maalej2017VANETsMA}
Y.~Maalej, S.~Sorour, A.~Abdel-Rahim, and M.~Guizani, ``Vanets meet autonomous
  vehicles: A multimodal 3d environment learning approach,'' in \emph{IEEE
  Global Communications Conference}, 2017, pp. 1--6.

\bibitem{arnold2020cooperative}
E.~Arnold, M.~Dianati, R.~de~Temple, and S.~Fallah, ``Cooperative perception
  for 3d object detection in driving scenarios using infrastructure sensors,''
  \emph{IEEE Transactions on Intelligent Transportation Systems}, 2020.

\bibitem{howe2021weakly}
M.~Howe, I.~Reid, and J.~Mackenzie, ``Weakly supervised training of monocular
  3d object detectors using wide baseline multi-view traffic camera data,'' in
  \emph{Brit. Mach. Vis. Conf.}, 2021.

\bibitem{yu2022dairv2x}
H.~Yu, Y.~Luo, M.~Shu, Y.~Huo, Z.~Yang, Y.~Shi, Z.~Guo, H.~Li, X.~Hu, J.~Yuan,
  and Z.~Nie, ``Dair-v2x: A large-scale dataset for vehicle-infrastructure
  cooperative 3d object detection,'' in \emph{Proceedings of the IEEE/CVF
  Conference on Computer Vision and Pattern Recognition}, 2022.

\bibitem{Manivasagam2020LiDARsimRL}
S.~Manivasagam, S.~Wang, K.~Wong, W.~Zeng, M.~Sazanovich, S.~Tan, B.~Yang,
  W.-C. Ma, and R.~Urtasun, ``Lidarsim: Realistic lidar simulation by
  leveraging the real world,'' in \emph{IEEE Conference on Computer Vision and
  Pattern Recognition (CVPR)}, 2020, pp. 11\,164--11\,173.

\bibitem{xu2021opencda}
R.~Xu, Y.~Guo, X.~Han, X.~Xia, H.~Xiang, and J.~Ma, ``Opencda: an open
  cooperative driving automation framework integrated with co-simulation,'' in
  \emph{IEEE International Intelligent Transportation Systems Conference
  (ITSC)}.\hskip 1em plus 0.5em minus 0.4em\relax IEEE, 2021, pp. 1155--1162.

\bibitem{arnold2021fastreg}
E.~Arnold, S.~Mozaffari, and M.~Dianati, ``Fast and robust registration of
  partially overlapping point clouds,'' \emph{IEEE Robotics and Automation
  Letters}, 2021.

\bibitem{Patil2019TheHD}
A.~Patil, S.~Malla, H.~Gang, and Y.-T. Chen, ``The h3d dataset for
  full-surround 3d multi-object detection and tracking in crowded urban
  scenes,'' \emph{2019 International Conference on Robotics and Automation
  (ICRA)}, pp. 9552--9557, 2019.

\bibitem{Pitropov2021CanadianAD}
M.~Pitropov, D.~Garcia, J.~Rebello, M.~Smart, C.~Wang, K.~Czarnecki, and S.~L.
  Waslander, ``Canadian adverse driving conditions dataset,'' \emph{The
  International Journal of Robotics Research}, vol.~40, pp. 681 -- 690, 2021.

\bibitem{Pham2020A3DDT}
Q.-H. Pham, P.~Sevestre, R.~Pahwa, H.~Zhan, C.~H. Pang, Y.~Chen, A.~Mustafa,
  V.~Chandrasekhar, and J.~Lin, ``A*3d dataset: Towards autonomous driving in
  challenging environments,'' in \emph{IEEE International Conference on
  Robotics and Automation (ICRA)}, 2020, pp. 2267--2273.

\bibitem{Yogamani2019WoodScapeAM}
S.~Yogamani, C.~Hughes, J.~Horgan, G.~Sistu, P.~Varley, D.~O'Dea, M.~Uriar,
  S.~Milz, M.~Simon, K.~Amende, C.~Witt, H.~Rashed, S.~Chennupati, S.~Nayak,
  S.~Mansoor, X.~Perroton, and P.~Perez, ``Woodscape: A multi-task,
  multi-camera fisheye dataset for autonomous driving,'' \emph{2019 IEEE/CVF
  International Conference on Computer Vision (ICCV)}, pp. 9307--9317, 2019.

\bibitem{Huang2018TheAD}
X.~Huang, X.~Cheng, Q.~Geng, B.~Cao, D.~Zhou, P.~Wang, Y.~Lin, and R.~Yang,
  ``The apolloscape dataset for autonomous driving,'' in \emph{IEEE/CVF
  Conference on Computer Vision and Pattern Recognition Workshops (CVPRW)},
  2018, pp. 1067--10\,676.

\bibitem{Cordts2016TheCD}
M.~Cordts, M.~Omran, S.~Ramos, T.~Rehfeld, M.~Enzweiler, R.~Benenson,
  U.~Franke, S.~Roth, and B.~Schiele, ``The cityscapes dataset for semantic
  urban scene understanding,'' in \emph{IEEE Conference on Computer Vision and
  Pattern Recognition (CVPR)}, 2016, pp. 3213--3223.

\bibitem{Ros2016TheSD}
G.~Ros, L.~Sellart, J.~Materzynska, D.~V{\'a}zquez, and A.~M. L{\'o}pez, ``The
  synthia dataset: A large collection of synthetic images for semantic
  segmentation of urban scenes,'' in \emph{IEEE Conference on Computer Vision
  and Pattern Recognition (CVPR)}, 2016, pp. 3234--3243.

\bibitem{Neuhold2017TheMV}
G.~Neuhold, T.~Ollmann, S.~R. Bul{\`o}, and P.~Kontschieder, ``The mapillary
  vistas dataset for semantic understanding of street scenes,'' in \emph{IEEE
  International Conference on Computer Vision (ICCV)}, 2017, pp. 5000--5009.

\bibitem{Behley2019SemanticKITTIAD}
J.~Behley, M.~Garbade, A.~Milioto, J.~Quenzel, S.~Behnke, C.~Stachniss, and
  J.~Gall, ``Semantickitti: A dataset for semantic scene understanding of lidar
  sequences,'' \emph{2019 IEEE/CVF International Conference on Computer Vision
  (ICCV)}, pp. 9296--9306, 2019.

\bibitem{Hu2020TowardsSS}
Q.~Hu, B.~Yang, S.~Khalid, W.~Xiao, A.~Trigoni, and A.~Markham, ``Towards
  semantic segmentation of urban-scale 3d point clouds: A dataset, benchmarks
  and challenges,'' in \emph{IEEE Conference on Computer Vision and Pattern
  Recognition (CVPR)}, 2021.

\bibitem{Chang2019Argoverse3T}
M.-F. Chang, J.~Lambert, P.~Sangkloy, J.~Singh, S.~Bak, A.~T. Hartnett,
  D.~Wang, P.~Carr, S.~Lucey, D.~Ramanan, and J.~Hays, ``Argoverse: 3d tracking
  and forecasting with rich maps,'' in \emph{Proc. IEEE Conf. Comput. Vis.
  Pattern Recognit.}, 2019, pp. 8740--8749.

\bibitem{Ettinger2021LargeSI}
S.~Ettinger, S.~Cheng, B.~Caine, C.~Liu, H.~Zhao, S.~Pradhan, Y.~Chai, B.~Sapp,
  C.~Qi, Y.~Zhou, Z.~Yang, A.~Chouard, P.~Sun, J.~Ngiam, V.~Vasudevan,
  A.~McCauley, J.~Shlens, and D.~Anguelov, ``Large scale interactive motion
  forecasting for autonomous driving : The waymo open motion dataset,''
  \emph{ArXiv}, vol. abs/2104.10133, 2021.

\bibitem{Jia2016EnhancedCC}
D.~Jia and D.~Ngoduy, ``Enhanced cooperative car-following traffic model with
  the combination of v2v and v2i communication,'' \emph{Transportation Research
  Part B-methodological}, vol.~90, pp. 172--191, 2016.

\bibitem{Kim2015MultivehicleCD}
S.-W. Kim, B.~Qin, Z.~J. Chong, X.~Shen, W.~Liu, M.~Ang, E.~Frazzoli, and
  D.~Rus, ``Multivehicle cooperative driving using cooperative perception:
  Design and experimental validation,'' \emph{IEEE Transactions on Intelligent
  Transportation Systems}, vol.~16, pp. 663--680, 2015.

\bibitem{Rawashdeh2018CollaborativeAD}
Z.~Y. Rawashdeh and Z.~Wang, ``Collaborative automated driving: A machine
  learning-based method to enhance the accuracy of shared information,'' in
  \emph{International Conference on Intelligent Transportation Systems (ITSC)},
  2018, pp. 3961--3966.

\bibitem{Chen2015DSRCAR}
Q.~Chen, T.~Yuan, J.~Hillenbrand, A.~Gern, T.~Roth, F.~Kuhnt, J.~M.
  Z{\"o}llner, J.~Breu, M.~Bogdanovic, and C.~Weiss, ``Dsrc and radar object
  matching for cooperative driver assistance systems,'' in \emph{IEEE
  Intelligent Vehicles Symposium (IV)}, 2015, pp. 1348--1354.

\bibitem{suo2021trafficsim}
S.~Suo, S.~Regalado, S.~Casas, and R.~Urtasun, ``Trafficsim: Learning to
  simulate realistic multi-agent behaviors,'' in \emph{Proceedings of the
  IEEE/CVF Conference on Computer Vision and Pattern Recognition}, 2021, pp.
  10\,400--10\,409.

\bibitem{tan2021scenegen}
S.~Tan, K.~Wong, S.~Wang, S.~Manivasagam, M.~Ren, and R.~Urtasun, ``Scenegen:
  Learning to generate realistic traffic scenes,'' in \emph{Proceedings of the
  IEEE/CVF Conference on Computer Vision and Pattern Recognition}, 2021, pp.
  892--901.

\bibitem{Wu2020MotionNetJP}
P.~Wu, S.~Chen, and D.~Metaxas, ``Motionnet: Joint perception and motion
  prediction for autonomous driving based on bird’s eye view maps,'' in
  \emph{Proc. IEEE Conf. Comput. Vis. Pattern Recognit.}, 2020, pp.
  11\,382--11\,392.

\bibitem{Luo2018FastAF}
W.~Luo, B.~Yang, and R.~Urtasun, ``Fast and furious: Real time end-to-end 3d
  detection, tracking and motion forecasting with a single convolutional net,''
  in \emph{Proc. IEEE Conf. Comput. Vis. Pattern Recognit.}, 2018, pp.
  3569--3577.

\bibitem{Luiten_2020}
J.~Luiten, A.~Osep, P.~Dendorfer, P.~Torr, A.~Geiger, L.~Leal-Taixé, and
  B.~Leibe, ``Hota: A higher order metric for evaluating multi-object
  tracking,'' \emph{International Journal of Computer Vision}, vol. 129, no.~2,
  p. 548–578, Oct 2020.

\bibitem{journals/ejivp/BernardinS08}
K.~Bernardin and R.~Stiefelhagen, ``Evaluating multiple object tracking
  performance: The clear mot metrics.'' \emph{EURASIP J. Image Video Process.},
  2008.

\bibitem{Bewley2016_sort}
A.~Bewley, Z.~Ge, L.~Ott, F.~Ramos, and B.~Upcroft, ``Simple online and
  realtime tracking,'' in \emph{2016 IEEE International Conference on Image
  Processing (ICIP)}, 2016, pp. 3464--3468.

\bibitem{Ronneberger2015}
O.~Ronneberger, P.~Fischer, and T.~Brox, ``{U-Net}: Convolutional networks for
  biomedical image segmentation,'' \emph{Medical Image Computing and
  Computer-Assisted Intervention – MICCAI 2015}, May 2015.

\end{thebibliography}

\end{document}